\documentclass[conference]{IEEEtran}
\IEEEoverridecommandlockouts

\usepackage{amsmath,amssymb,amsfonts}
\usepackage{algorithmic}
\usepackage{booktabs}
\usepackage{cite}
\usepackage{url}
\usepackage{graphicx}
\usepackage{gensymb}
\usepackage{subcaption}

\usepackage{textcomp}
\usepackage{xcolor}
\def\BibTeX{{\rm B\kern-.05em{\sc i\kern-.025em b}\kern-.08em
    T\kern-.1667em\lower.7ex\hbox{E}\kern-.125emX}}
\begin{document}

\title{CAR-Net: A Cascade Refinement Network for \\ Rotational Motion Deblurring under \\ Angle Information Uncertainty
\thanks{This work was supported by JSPS KAKENHI Grant No. JP23K03913.}
}

\author{\IEEEauthorblockN{KA CHUNG LAI}
\IEEEauthorblockA{\textit{The Hong Kong Polytechnic University} \\
Hong Kong SAR, China \\
22080062d@connect.polyu.hk}
\and
\IEEEauthorblockN{AHMET CETINKAYA}
\IEEEauthorblockA{\textit{Shibaura Institute of Technology}\\
Tokyo, Japan \\
ahmet@shibaura-it.ac.jp}
}

\maketitle

\begin{abstract}

We propose a new neural network architecture called CAR-net (CAscade Refinement Network) to deblur images that are subject to rotational motion blur. Our architecture is specifically designed for the semi-blind scenarios where only noisy information of the rotational motion blur angle is available. The core of our approach is progressive refinement process that starts with an initial deblurred estimate obtained from frequency-domain inversion; A series of refinement stages take the current deblurred image to predict and apply residual correction to the current estimate, progressively suppressing artifacts and restoring fine details. To handle parameter uncertainty, our architecture accommodates an optional angle detection module which can be trained end-to-end with refinement modules. We provide a detailed description of our architecture and illustrate its efficiency through experiments using both synthetic and real-life images. Our code and model as well as the links to the datasets are available at \url{https://github.com/tony123105/CAR-Net}
\end{abstract}


\section{Introduction}
Image clarity is a fundamental prerequisite for numerous computer vision task, spanning domains from object recognition and autonomous navigation to medical imaging and forensic analysis. Motion blur, one of the most destructive forms of image degradation, comes from relative motion between the camera and scene deblurring exposure. This artifact can blur fine details, destroy critical information, and significantly degrade the performance of downstream algorithms.

Among various type of motion blur, Rotational Motion Blur (RMB) presents particular challenges. Unlike uniform linear motion, RMD generates spatially-variant Point Spread Function, where blur intensity and direction vary with pixel's distance from the rotational center analyzed by Sawchuk\cite{sawchuk1972}. This non-uniform nature makes it difficult for traditional deconvolution methods or deep learning model to effectively restore the sharp image.

A critical and practical challenge is the semi-blind scenario, where the initial blur angle estimation (derived from sensors) inherently contain noise. This presents a significant challenge to existing state-of-the-art methods. For instance, recent advanced framework such as PCS-RMD\cite{qin2025} have achieved remarkable result, but they operate under idealized non-blind assumption, where the ground truth angle is assumed fully known.

Consequently, these state-of-art models are fundamentally not designed to handle the parameter uncertainties common in real-world applications. Their performance degrade significantly under noisy angle inputs, as they lack dedicated mechanisms to correct these physical parameter errors. This reveals a critical gap between current academic solutions and practical application requirements.

To address this gap, we propose CAR-Net (CAscade Refinement Network), a hybrid framework that incocproates a combination of physics-based modeling and deep learning. Our method first performs an initial deblurring in the Polar Coordinate System based on noisy angle. The core innovation lies in the Angle Detection module (AD), which learns to predict the ground truth angle by analyzing artifacts generated during first deblurring. These optimized angles are then used for more precise deblurring, and the output is processed through a Progressive Spatial Refinement Network to recover fine details.
Our main contributions are as follows:
\begin{itemize}
    \item We propose a novel multi-stage hybrid architecture that effectively combines physics-based frequency-domain inversion with a progressive deep learning refinement network, specifically optimized for rotational blur.
    \item We introduce a dedicated angle detection module that learns to correct inaccuracies in the initial blur angle estimate by creating a corrective feedback loop, thus addressing semi-blind deblurring.
    \item We demonstrate through extensive experiments to validate that the complete CAR-Net-AD is robust to significant parameter uncertainty, outperforming existing state-of-the-art methods in both quantitative metrics and visual quality while being more computational efficient.
\end{itemize}
The rest of the paper is organized as follows. In Section~\ref{Sec:RelatedWork}, we review prior work in related fields. Section~\ref{Sec:Problem_Fromulation}, we present the problem formulation, while in Section~\ref{Sec:Methodology} we provide details of the architecture of the proposed method; In Section~\ref{Sec:Experiment_setup}, we describe our experimental setup, followed by a comprehensive presentation and analysis of our result in Section~\ref{Sec:Result}. Finally in Section~\ref{Sec:Discussion}, we conclude the paper and provide a list of future works related to extending our model.

\section{Related Work on Rotational Deblurring} 
\label{Sec:RelatedWork}

Our work is related to four important classes of image-deblurring strategies: 1) methods that operate in polar coordinates, 2) progressive and hybrid deblurring frameworks, 3) deep unrolling frameworks that combine physics and learning, and 4) approaches for handling inaccurate blur parameters.

\subsection{Rotational Deblurring in the Polar Coordinate System}

The strategy of transforming the rotational blurred image into Polar Coordinate System (PCS) helps simplify the deblurring problem. This approach was described in the foundational work of Sawchuk \cite{sawchuk1972}, who demonstrated that rotational blur that is complex and spatially-variant in the Cartesian domain becomes more tractable and \emph{spatially-invariant} along the angular axis of PCS. 

Several approaches have been proposed to utilize this strategy of transformation to polar coordinates. For instance, Morimoto et al. \cite{morimoto2011restoration} demonstrated deblurring by applying inverse filtering in the PCS. More recently, Krouglova et al. \cite{krouglova2022restoration} employed a similar PCS-based strategy, but utilized Wiener filter for encountering noise and artifacts. In our approach, we have a module (named InversionModule) that follows this strategy of transforming the image into PCS and performing frequency-domain deconvolution and transforming it back to Cartesian domain. This module is used in the initial stage of our framework.

Transformation from Cartesian coordinates to polar coordinates causes some artifacts, because the necessary resampling and interpolation processes during the Cartesian-to-Polar mapping degrade certain elements (high-frequency details) in the image \cite{qin2025}. To resolve such artifacts, Qin et al.~\cite{qin2025} recently proposed the integration of deep learning into the PCS framework. In particular, they introduced a deep learning-based error correction module for Cartesian to polar coordinate transformation. In our work, we propose to address such artifacts using a different mechanism via a progressive spatial-refinement module. 

\subsection{Progressive and Hybrid Deblurring Frameworks}
Recent work of Qin et al.~\cite{qin2024progressive} has demonstrated significant advantages of a progressive multi-stage framework that combines a model-based coarse deblurring step with a learning-based refinement step. Their progressive framework for rotational motion deblurring proposed stands as a current state-of-the-art example. Their method first employs a sophisticated model-based deconvolution and then uses a deep refinement network (TDM-CNN) to reduce artifacts.

While our work shares this high-level coarse-to-fine philosophy, our underlying mechanisms and core contributions are fundamentally distinct in two key ways.

First, our coarse deblurring stage applies 2D deconvolution directly on the image in PCS. In contrast, the approach of Qin et al.~\cite{qin2024progressive} is based on the Blurring Paths Method (BPM), which decomposes the 2D image into a series of 1D sequences for processing.

Second, our framework introduces a novel feedback loop to handle uncertainty in the blur angle information. In Qin et al.~\cite{qin2024progressive}, the blur angle information is assumed to be noise-free. More specifically, they introduce a "blur extents factor" to adaptively tune the regularization of their deconvolution which assumes the input blur angle is accurate. Our work addresses a more challenging semi-blind problem by integrating an Angle Detection Module which is a CNN that learns to predict and correct the errors of initial angle estimate. Thus, our proposed architecture CAR-Net offers fundamentally different approach to robust deblurring.

\subsection{Deep Unrolling for Image Deblurring}

Deep unrolling methods involve mapping the iteration of a classical optimization algorithm into a fixed-depth neural network, which have become a powerful paradigm for creating hybrid, physics-guided networks~\cite{gregor2010learning}. Pioneering work in image restoration, such as Zhang et al. \cite{zhang2017learning} demonstrated the strength of this approach by embedding pre-trained CNN denoisers as learning priors within iterative optimization schemes. The PCS-based approach of Qin et al. \cite{qin2025} mentioned previously, is a concrete application of this powerful paradigm.

Although, our model is also iterative, alternating between a physics-based step (inversion) and a learner-based step (refinement), it differs from the strict deep unrolling paradigm. Our model employs a cascade of corrective modules with distinct sub-goals: first, the Angle Detection Module aims to find the ground truth (correct blur angle); second, modules in the Refinement Stage perform image enhancements. These offer more a flexible approach than a formal optimization unrolling. This decouples parameter correction from image refinement which allows each specialized module to learn a simpler, more constrained task without being overloaded by the demands of the single, complex objective function.

\subsection{Handling Noisy or Inaccurate Blur Parameters}

The core innovation of our work lies in addressing the “semi-blind” problem, where the provided blur angle information is an imperfect estimate of the true value, which results in an inaccurate blur kernel. This represents a common and practical real-world scenario.

The critical importance of this problem is highlighted by classical PCS-based restoration methods, where, as noted by Krouglova et al.\cite{krouglova2022restoration}, even small errors are amplified by the blur. This sensitivity motivates the need for methods that are robust to parameter uncertainty. This sensitivity motivates the need for methods that are robust to parameter uncertainty.

This challenge exists within the the broader context of robust deblurring. For instance, Nan and Ji's work \cite{nan2020deep} employs an error-in-variables model to address general blur kernel imprecision. While powerful, such general-purpose methods are not explicitly tailored to the parametric, spatially-variant nature of rotational blur.

In contrast, our work is more direct and it provide a data-driven solution focused specifically on the domain of rotational motion deblurring. The Angle Detection Module in our proposed in architecture is a CNN that is specially trained to recognize the unique artifact patterns caused by angular errors in RMD. It performs precise regression corrections targeting for \emph{single physical parameter} (blur angle). This learning-based parameter correction method differs fundamentally from more general kernel function correction models.

\section{Problem Formulation}
\label{Sec:Problem_Fromulation}
\subsection{Rotational Blur Problem}
Image blur is typically considered as a degradation process that can be described as follows: the latent sharp image $f$ is convolved with the Point Spread Function (PSF), followed by the noise \cite{Gonzalez2018}. For the specific case of rotational motion, the blurred image observed in the Cartesian coordinate system $g_\mathrm{c}$ is characterized as

\begin{equation}
    g_\mathrm{c} = f_\mathrm{c} \otimes K(\theta_{\mathrm{GT}},c) + n
    \label{eq:degradation_model_cartesian}
\end{equation}
where $f_\mathrm{c}$ is the latent sharp image, $\otimes$ denotes the 2D convolution operator, $n$ is additive noise, and $K$ is the spatially-variant Point Spread Function (PSF). The PSF is parameterized by the ground-truth rotation angle $\theta_{\mathrm{GT}}$, and rotation center ($c=\left(c_\mathrm{x},c_\mathrm{y}\right)$). 

To simplify the spatially-variant problem, we adopt the strategy, first proposed by Sawchuk \cite{sawchuk1972}, of transforming the image into the Polar Coordinate System (PCS). In the PCS, the degradation model becomes

\begin{equation}
    g_\mathrm{p} = f_\mathrm{p} * K_\mathrm{p}(\theta_{\mathrm{GT}},c) + n_\mathrm{p}
    \label{eq:degradation_model_polar}
\end{equation}
where $g_\mathrm{p}$, $f_\mathrm{p}$ and $n_\mathrm{p}$ are the PCS representation of the blurred image, sharp image, and noise respectively.

Crucially, as described in Section~\ref{Sec:RelatedWork} of \cite{sawchuk1972} the convolution kernel $K_\mathrm{p}$ is simplified to a spatially-invariant kernel, since the complex rotational motion is simplified to a one-dimensional blurring effect that acts only along the angular axis. Thus, $K_\mathrm{p}$ is a simpler function dependent solely on the blur angle $\theta_{\mathrm{GT}}$. Furthermore, the convolution operator * in Eq. \eqref{eq:degradation_model_polar} is a circular convolution along the the angular dimension. This simplification of degradation model is the fundamental motivation for operating in the PCS and is foundational to our entire deconvolution approach.

\subsection{Challenge of Parameter Inaccuracy}
For non-blind deblurring scenarios, the ground-truth angle $\theta_\mathrm{GT}$ is perfectly known. However, our semi-blind setting assumes that an initial estimate, $\theta_\mathrm{initial}$ is provided which is close but imperfect. This practical scenario can be modeled as the true angle corrupted by some unknown error, $\epsilon$, as given by
\begin{equation}
    \theta_\mathrm{initial} = \theta_\mathrm{GT} + \epsilon.
    \label{eq:angle_error}
\end{equation}

Here, the core challenge lies in the fact that using a deconvolution kernel base on the imprecise angle ($K(\theta_\mathrm{initial})$), leads to severe mismatch with the actual degradation process. This mismatch results in suboptimal restoration quality, characterized by significant artifacts. Our primary objective is to design a framework that can estimate and correct the angular error ($\epsilon$) to perform the final image restoration.

\section{Methodology}
\label{Sec:Methodology}
Our proposed framework, which we named CAR-Net (Cascade Refinement Network), is a multi-stage model designed to restore rotational blurred images progressively in PCS.

We present our method in two distinct configurations:

\begin{enumerate}
    \item Our baseline hybrid Inversion-Refinement model, which combines classical deconvolution with deep refinement
    \item Our full model, which incorporates a novel Angle Detection Module to create a corrective feedback loop, making the framework robust to parameter inaccuracies.
\end{enumerate}

The initial step for both configurations involves transforming the input blurred image, $g_\mathrm{c}$, via a Cartesian-to-Polar Transformation (CPT) to produce its polar representation $g_\mathrm{p}$. All subsequent core modules operate on this polar representation.

\subsection{Baseline Model: Hybrid Inversion-Refinement}
\begin{figure}[t]
    \centering
    \includegraphics[width=\columnwidth]{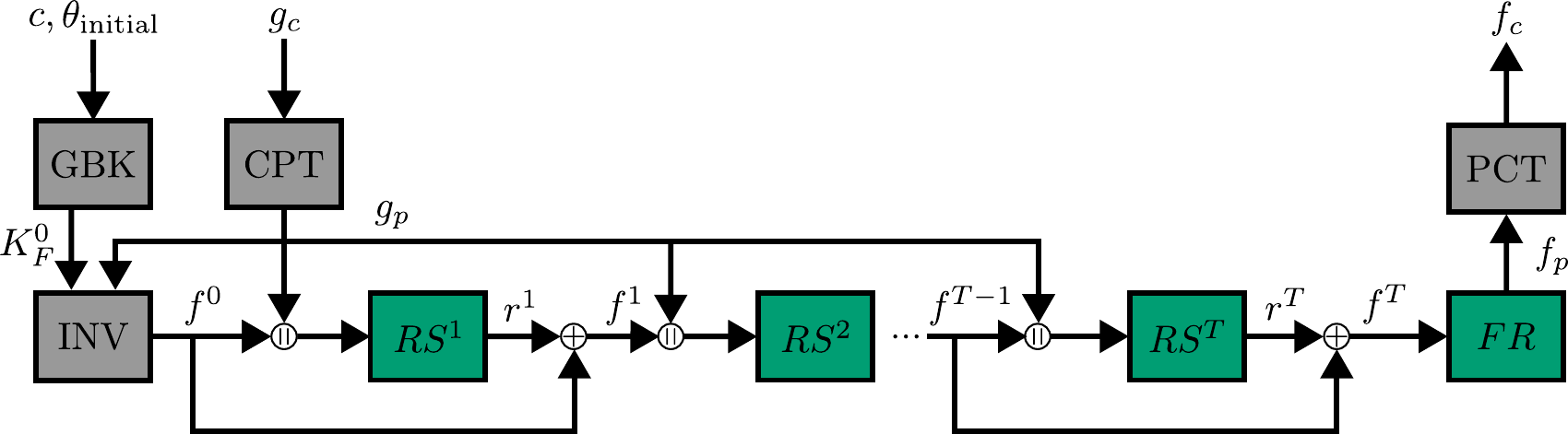}
    \caption{The architecture of our \emph{Baseline Model}. The input polar image $g_\mathrm{p}$ is first processed by the Frequency Inversion Module (INV) to produce an initial estimate $f_0$. This estimate is then passed to the Progressive Spatial Refinement Module (RS), which iteratively remove artifacts and restores details. Green and gray blocks denote trainable and non-trainable modules, respectively.}
    \label{fig:arch_baseline}
\end{figure}

Our baseline model builds a powerful two-step pipeline that serves a strong foundation for rotational deblurring. The complete data flow of this model is illustrated in Fig~\ref{fig:arch_baseline}.

\paragraph{Step 1: Frequency Inversion Module}
The process begins with an initial deblurring in frequency domain. This provides a computationally efficient initial estimate of the sharp image by inverting the blur process. It is based on convolution theorem \cite{jahne2005digital}, which states that an element-wise multiplication in the frequency domain is equivalent to convolution in the spatial domain.

In Frequency Inversion Module, we apply 2D Fast Fourier Transform (FFT) and its inverse ($\mathcal{F}^{-1}$) to get the equation
\begin{equation}
f = \Re\left\{ \mathcal{F}^{-1}\left( \frac{\mathcal{F}(g_\mathrm{p})}{K_\mathrm{F} + \epsilon} \right) \right\},
\label{eq:inversion}
\end{equation}
where 
$\mathcal{F}$ and $\mathcal{F}^{-1}$ denote 2D FFT and its inverse,
$K_\mathrm{F}$ is pre-computed 2D FFT of blur kernel (PSF),
$\Re$ denotes the operation of taking the real part of the complex result,
$\epsilon$ is small regularization constant for numerical stability ($10^{-8}$)
This step effectively removes the global blur, but often leaves behind artifacts.

\paragraph{Step 2: Progressive Spatial Refinement Module (RM)}

This module is a multi-stage network that progressively refines the initial estimate, removing artifacts such as ringing and noise amplification.

The architecture of this module is illustrated in Fig.~\ref{fig:arch_baseline}. The initial estimate $f^0$ computed in the previous step is passed to this module. To break down the complex deblurring operation into sequence of simpler corrective steps, this module contains a series of structurally identical residual refinement stages with their own blocks ($RS^k$) followed by a final refinement ($FR$) block. We consider $T$ number of refinement stages (i.e., $RS^1,RS^2,\ldots ,RS^T$). The detailed architectures of these blocks are shown in Figs.~\ref{fig:refinement_stage} and \ref{fig:final_refinement}. 

At each stage $RS^k$, the network takes the current deblurred estimate $f^{k-1}$ and the original blurred image $g_\mathrm{p}$ as inputs to predict a residual image, which is then added to the current estimate to produce the next iteration image $f^k$. The process will be updated based on the formula 
\begin{equation}
f^{k} = f^{k-1} + r^{k} = f^{k-1} + {RS}^k(f^{k-1} \mathbin{\|} g_\mathrm{p})
\label{eq:progressive_residual}
\end{equation}
where $r^k$ is the residue from stage $k$ and
$\mathbin{\|}$ denotes channel-wise concatenation of two tensors.

After the final refinement stage, the restored polar image is transformed back to Cartesian coordinate via a Polar-to-Cartesian Transformation (PCT).

\subsection{Full Model with Angle Correction}
\begin{figure}[tbp]
    \centering
    \includegraphics[width=\columnwidth]{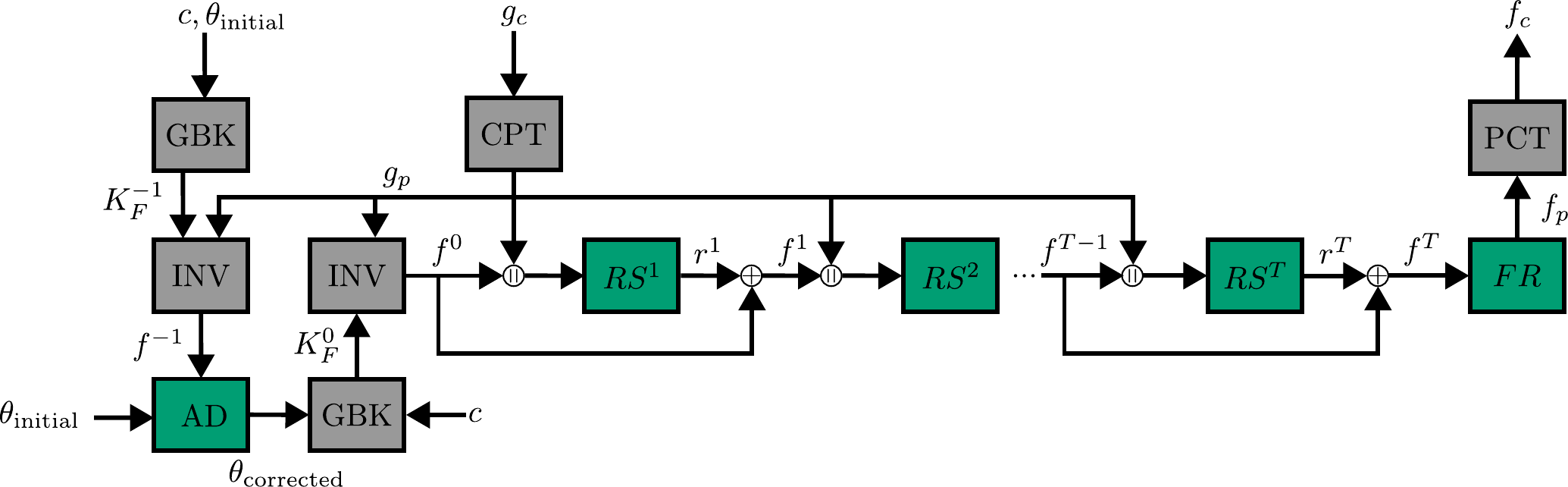}
    \caption{The architecture of our \emph{Full Model} which extends the baseline by inserting the Angle Detection Module (AD) into a feedback loop. The AD accept the initial deblurred image ($f^{-1}$) and the initial angle ($\theta_\mathrm{initial}$) to predict a corrected angle ($\theta_\mathrm{corrected}$). The new angle is used to perform a more accurate inversion before the final refinement stages. Green and gray blocks denote trainable and non-trainable modules, respectively.}
    \label{fig:arch_full}
\end{figure}

The performance of the baseline model is highly dependent on the accuracy of the initial blur angle. Our full model addresses this limitation by introducing a novel Angle Detection Module (AD). This enhanced architecture is depicted in Fig~\ref{fig:arch_full}.

\begin{figure*}[t]
    \centering
    \begin{subfigure}[b]{0.2\textwidth}
        \centering
        \includegraphics[width=\linewidth]{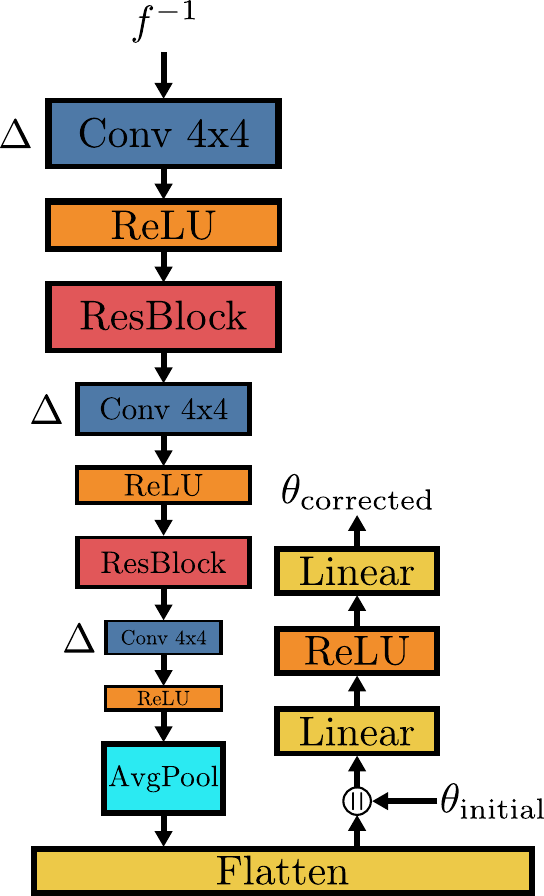}
        \caption{Angle Detection Module}
        \label{fig:angle_detection}
    \end{subfigure}
    \hfill
    \begin{subfigure}[b]{0.18\textwidth}
        \centering
        \includegraphics[width=\linewidth]{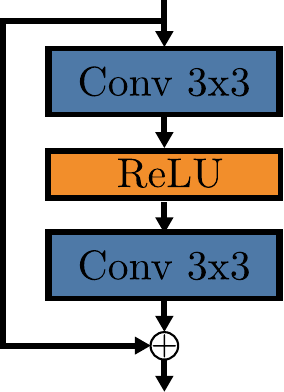}
        \caption{ResBlock}
        \label{fig:resblock}
    \end{subfigure}
    \hfill
    \begin{subfigure}[b]{0.18\textwidth}
        \centering
        \includegraphics[width=\linewidth]{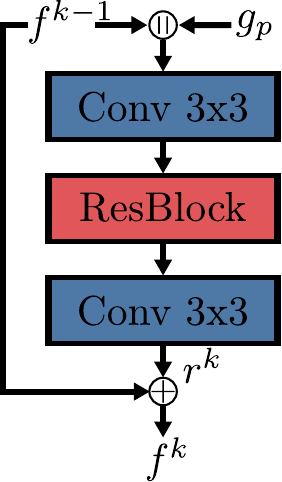}
        \caption{Refinement Stage}
        \label{fig:refinement_stage}
    \end{subfigure}
    \hfill
    \begin{subfigure}[b]{0.18\textwidth}
        \centering
        \includegraphics[width=\linewidth]{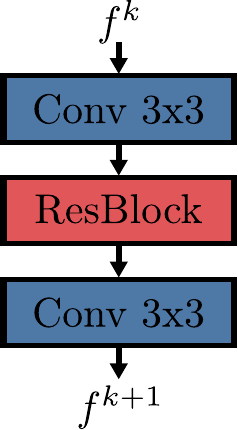}
        \caption{Final Refinement}
        \label{fig:final_refinement}
    \end{subfigure}
    \caption{The architectural components of our proposed model. (a) The module for correcting the blur angle. (b) The residual block. (c) The iterative refinement stage. (d) The final refinement block. Module shown with $\Delta$ represents where the feature map is downsampled}
    \label{fig:detail_arch}
\end{figure*}

\paragraph{Angle Detection Module (AD)}
To address the semi-blind problem, the Angle Detection Module (AD) is a dedicated CNN trained to predict a highly accurate angle estimate. It learns to recognize the specific artifact patterns present in the initial deblurred image (identified in the full model as $f^{-1}$) that result from the initial angle guess. Using both the image artifacts and the initial angle ($\theta_\mathrm{initial}$) as input, it directly regresses the final corrected angle, $\theta_\mathrm{corrected}$ as
\begin{equation}
\theta_\mathrm{corrected} = \mathrm{AngleDetector}(f^\mathrm{-1},\theta_\mathrm{initial}) 
\end{equation}
The detailed architecture of AD is shown in Fig.~\ref{fig:angle_detection}. In this architecture, a feature extraction backbone processes the input $f^{-1}$, and the output is compressed into a feature vector through global average pooling. This vector is concatenated with $\theta_\mathrm{initial}$, and a final pass through the multilayer perceptron regresses the corrected angle.

\subsection{Loss function}
To train the network effectively, 4 different loss functions are combined for enforcing pixel-level accuracy, perceptual similarity, and physical consistency. The loss is calculated on the output of each stage to produce a better prediction of the sharp image progressively.

Specifically, the total loss $\mathcal{L}_\mathrm{total}$ is defined by
\begin{align}
\label{eq:combined_loss}
\mathcal{L}_\mathrm{total} 
={}& w_\mathrm{L1}\mathcal{L}_\mathrm{L1} 
+ w_\mathrm{SSIM} \mathcal{L}_\mathrm{SSIM} \notag\\
&+ w_\mathrm{physics}\mathcal{L}_\mathrm{physics}
+ w_\mathrm{angle}\mathcal{L}_\mathrm{angle}
\end{align}
where ${w_{\mathrm{physics}}}$, ${w_{\mathrm{SSIM}}}$, ${w_{\mathrm{L1}}}$, and $w_\mathrm{angle}$ are the weights for the respective loss components, which we describe below.

\subsubsection{L1 Reconstruction Loss}
To enforce pixel-level accuracy and avoid outliers from the noise, we compute the L1 norm (Mean Absolute Error) between the model's deblurred output and the ground-truth image. 

The loss averaged over all refinement stages (including the final one) is given by
\begin{equation}
\label{eq:L1_loss}
\mathcal{L}_\mathrm{L1} = \frac{1}{T+1} \sum_{k=1}^{T+1} \frac{1}{N} \|f^{k} - f_{\mathrm{sharp}}\|_1,
\end{equation}
where $f^k$ is the output of the refinement block $RS^k$ for $k\in\{1,2,\ldots, T\}$, $f^{T+1}=f_p$ is the output of the final refinement block $FR$, and $N$ is the number of pixels in the image.

\subsubsection{Structural Similarity Index (SSIM) loss}
To ensure perceptual and structural quality, Structural Similarity Index (SSIM) loss \cite{wang2004image} is used. The SSIM loss can figure out the structural information, like luminance, contrast, and edges, other than L1 losses. Since a larger value of SSIM is preferable, to ensure backpropagation works properly, $1-\mathrm{SSIM}$ is used for calculating the loss and it is averaged over all the refinement stages (including the final one) by
\begin{equation}
\label{eq:SSIM_loss}
\mathcal{L}_{\mathrm{SSIM}} = \frac{1}{T+1} \sum_{k=1}^{T+1} (1 - \mathrm{SSIM}(f^{k}, f_{\mathrm{sharp}})),
\end{equation}
where $f^k$ is the output of the refinement block $RS^k$ for $k\in\{1,2,\ldots, T\}$, and $f^{T+1}=f_p$ is the output of the final refinement block $FR$. 

\subsubsection{Physics-based Consistency Loss}
To enforce physical realism, we incorporate a Physics-based Consistency Loss that ensures the deblurring process is a valid inverse of the blur formation \cite{nah2021clean}. Our method simulates the rotational blur by performing a backward mapping for each pixel in the deblurred output. This is achieved by calculating the coordinates of $n_{\mathrm{step}}$
discrete points along the pixel's rotational arc, sampling their values via bilinear interpolation, and averaging them to produce the final re-blurred result. This re-blurred image is then compared to the original blurred input image $g_\mathrm{p}$ using an L1 loss using the formula
\begin{equation}
\label{eq:physical_loss}
\mathcal{L}_{\mathrm{physics}} = \frac{1}{T+1} \sum_{k=1}^{T+1} \|\mathrm{Blur}(f^{k}, \theta_{\mathrm{corrected}}) - g_\mathrm{p}\|_1,
\end{equation}
where $f^k$  is the output of refinement block $RS^k$ for $k\in\{1,2,\ldots, T\}$, and $f^{T+1}=f_p$ is the output of the final refinement block $FR$. 

\subsubsection{Angle Regression Loss}
For our Full model, to encourage the model estimate the blur angle accurately, we use normalized L1 loss between the predicted angle $\theta_{\mathrm{corrected}}$ and the ground-truth angle $\theta_{\mathrm{GT}}$. Both angles are normalized by the maximum possible angle $\theta_{\mathrm{max}}$ to scale the loss to a consistent range, improving training stability by
\begin{equation}
\label{eq:angle_loss}
\mathcal{L}_{\mathrm{angle}} = \left\| \frac{\theta_{\mathrm{corrected}}}{\theta_{\mathrm{max}}} - \frac{\theta_{\mathrm{GT}}}{\theta_{\mathrm{max}}} \right\|_1=\frac{|\theta_{\mathrm{corrected}}-\theta_{\mathrm {GT}}|}{\theta_{\mathrm{max}}}.
\end{equation}
Note that this angle regression loss is applied only when training our full CAR-Net-AD. It is excluded when training our base model, as the base model doesn't contain angle detection module and performs no angle correction.

\section{Experiment Setup}
In this section, we explain our datasets and the setup for our experiments.
\label{Sec:Experiment_setup}
\subsection{Dataset}
In our experiments, we use both simple pattern-based images as well as real-world images as datasets.  
\subsubsection{Simple Pattern Rotational Motion Blur Datasets}
We consider a set of simple geometric patterns in images of size $320\times 320$. These images may contains 11 different patterns (checkerboard, circles, diamond, grid, l shape, line, radial, sine grating, spiral, star, triangle). In particular, for each pattern, 10 images are generated. 

The rotational blur to each sharp image is applied by using the GNU Image Manipulation Program (GIMP) \cite{GIMP2025}. The blur is generated with a fixed central rotational point and a discrete set of blur angles, $\theta \in \{1\degree, 2\degree,\dots,40\degree\}$. This process results in a training set of $\mathrm{11}\times\mathrm{10}\times\mathrm{40} =\mathrm{4,400}$ blurred-sharp image pairs.

\subsubsection{Real-World Rotational Motion Blur Datasets}
The primary training and testing data is derived from the real-world rotational motion blur dataset introduced by Qin et al.\cite{qin2024progressive}. This dataset provides high-quality sharp images of 15 different scenes captured at slightly angle difference.

\paragraph{Training Data}
From the provided sharp images, we create a base set of 101 sharp images for training. This was done by selecting 7 distinct angular poses for 14 set of scenes, and 3 poses for the remaining scene. We then apply rotational blur to each of these sharp image using GIMP's circular motion blur filter. The blur was generated with the following parameters:

\begin{itemize}
    \item \textbf{Rotational Center:} Fixed at the image center (0.5,0.5).
    \item \textbf{Blur Angle:} A discrete set of 40 integer angles, $\theta \in \{1\degree,2\degree,\dots,40\degree\}$
    \item \textbf{Blur Amount:} A parameter corresponding to the number of motion steps in GIMP, randomly sampled from  $\{{1,2,\dots,15}\}$.
\end{itemize}
This process yielded a training set of 101 sharp images $\times$ 40 angles = $\mathrm{4,040}$ blurred-sharp pairs.

\paragraph{Testing Data}
For testing, we used test set of scenes provided by Qin et al. \cite{qin2024progressive}. To ensure a fair and challenging evaluation of our model's generalization capabilities, the test blur were generated using GIMP circular motion blur filter with continuous random angle uniformly sampled from the range $[1.0, 40.0]$ degrees. This ensures that our model is evaluated on its ability to deblur angle it has not explicitly seen during training.

\subsection{Simulation of Angle Inaccuracy}
To verify our angle detection module is robust to inaccuracies in initial angle estimate, we designed a specific set of experiments for the parameter inaccuracy. Given that our dataset provides the ground-truth angle, $\theta_\mathrm{{GT}}$, for every blurred image, we generate the imperfect angle (which is the input to our model) by adding zero-mean Gaussian noise as
\begin{equation}
    \theta_\mathrm{initial} = \theta_\mathrm{GT} + \epsilon,
    \label{eq:noise_simulation}
\end{equation}
where $\epsilon \sim \mathcal{N}(0,\sigma^2)$.
The standard deviation, $\sigma$, is a hyperparameter we vary in our ablation studies to simulate different levels of uncertainty.

\subsection{Implementation Details}
All models were implemented using the PyTorch framework \cite{paszke2019pytorch}. To optimize the network parameters, Adam optimizer \cite{kingma2014adam} was used with an initial learning rate of $\mathrm{1x10^{-4}}$.

To manage the learning rate during training, ReduceLROnPlateau scheduler was employed. This scheduler monitors the validation loss and reduces the learning rate by half if the loss does not improve for 5 consecutive epochs. This strategy permits for larger learning steps early in the training process and fine adjustments as the model converges to a solution.

Our models were trained for 200 epochs with batch size of 4. For the Refinement module, the number of iterative stages are set to $T=3$. The weight of our loss function were set to $w_\mathrm{L1}=1.0$, $w_\mathrm{SSIM}=0.5$, $w_\mathrm{physics}=0.1$, and $w_\mathrm{angle}=0.1$. For the physical-based consistency loss, the number of discrete sampling points, $n_{\mathrm{step}}$, was set to 15. All the experiments were conducted in a single NIVIDA RTX 3080ti GPU.

\subsection{Evaluation Metrics}
To quantitatively evaluate and compare the performance of all models, we employed two standard and complementary image quality assessment metrics. Both metrics are computed by comparing the deblurred output image against the ground-truth sharp image.

\begin{itemize}
    \item \textbf{Peak Signal-to-Noise-Ratio (PSNR):} PSNR is the most commonly used metric for measuring pixel-level reconstruction fidelity. It is based on the Mean Squared Error (MSE) between the restored and ground-truth images. It is reported in decibels (dB), and a higher PSNR value indicates a more accurate reconstruction.
    \item \textbf{Structural Similarity Index (SSIM)\cite{wang2004image}:} In contrast to PSNR, SSIM is designed to measure perceptual quality by comparing structural information, luminance, and contrast between two images. The SSIM value ranges from -1 to 1, where 1 signifies a perfect structural match.
\end{itemize}

\subsubsection{Evaluation Protocol for PCS-based Methods}
The key factor in the design of our evaluation protocol is to ensure fair and accurate comparison. We note that the Polar-to-Cartesian Transformation (PCT), serving as the essential final step for all models operating in polar coordinates, produces Cartesian images where corner regions may become undefined or filled with artifacts due to exceeding the polar grid's boundaries. Furthermore, these corner regions remain unaffected by central rotational blur. To confine the evaluation scope to the effective deblurring region, all metrics (PSNR and SSIM) are computed only within the largest circular region of interest (ROI) that is inscribed within the square output image. This approach ensures that the reported scores are true reflections of the core deblurring algorithm's restoration quality, unaffected by the transformation artifacts, thus providing a fair comparative benchmark for all PCS-based methods.

\subsection{Baselines for Comparison}
We aim to compare our proposed models against the representative stat-of-the-art model from the literature. In particular, we consider the following list of models and methods:
\begin{itemize}
    \item RMD\_PCS\cite{qin2025}, 
    \item Our baseline model, CAR-Net-Base,
    \item Our full model, CAR-Net-AD, which includes the Angle Detection Module.
\end{itemize}
This comparison enables us to evaluate the benefit of our refinement network, the advantage of our dedicated architecture, and the performance gain provided by the novel angle correction mechanism.

\section{Results and Analysis}
\label{Sec:Result}
In this section, we present a comprehensive evaluation of our proposed models, CAR-Net-Base and CAR-Net-AD. We begin our evaluation with a series of in-depth ablation studies to validate our architectural choices. We then conclude with quantitative comparison against a state-of-the-art baseline to demonstrate the superiority of our final, optimized model. Unless otherwise specified, all models are considered to be trained on our real-world dataset using a noise strategy ($\sigma=5$) which we mentioned in Section~\ref{Sec:Experiment_setup} and evaluated on the real-world dataset under noise angle $\sigma=5$ test set.

\subsection{Ablation Studies}

\subsubsection{Effectiveness of Proposed Modules}
To validate the effectiveness of the core architecture selection, we first analyze the performance contribution of each proposed module. We use an inversion module as a non-learned baseline, we progressively incorporate deep learning components. As shown in Table~\ref{tab:ablation_modules}, applying the progressive residual inversion module (forming the base CAR-Net) significantly improves PSNR by 1.4 dB, demonstrating its powerful capability in eliminating deconvolution artifacts. Subsequently, incorporating the angle detection module (forming the complete CAR-Net-AD) delivers further improvement, confirming that clearly correcting the blurring angle before the refinement is an extremely effective strategy. However, applying angle detection module in isolation leads to a slight decrease in PSNR by 0.08, which highlights that the primary strength of angle detection module lies in its synergy with the refinement network; by correcting the blur angle, it provides a cleaner input that enables Refinement Module to restore final details much more effectively.

\begin{table}[tbp]
\centering
\caption{Experiment on contribution of the Angle Detection module (AD) and Refinement Module (RM)}
\label{tab:ablation_modules}
\begin{tabular}{cc cc}
\toprule
\textbf{AD} & \textbf{RM} & \textbf{PSNR (dB)} & \textbf{SSIM}\\
\midrule
& &  22.43 & 0.7780 \\
\checkmark & & 22.35 & 0.7779 \\
 & \checkmark& 23.84 & \textbf{0.8034} \\
\checkmark & \checkmark & \textbf{24.03} & 0.8030 \\
\bottomrule
\end{tabular}
\end{table}

\subsubsection{Impact of Number of Refinement Stages}
We analyzed the effect of $T$, the number of $RS^k$ stages in the refinement module, to determine the optimal depth. As shown in Table~\ref{tab:stage}, performance improves as the number of stages increase from one to three. However, a slight performance degradation is observed when the fourth stage is added. This suggests that three stages provide the most optimal balance between refinement capacity and model complexity for this task. Based on this result, all subsequent experiments use a \textbf{3-stage} configuration of our models.

\begin{table}[tbp]
\centering
\caption{Ablation Study on the result of Number of stages in Refinement Module}
\label{tab:stage}
\begin{tabular}{c cc}
\toprule
\textbf{Number of Stages ($T$)} & \textbf{PSNR (dB)} & \textbf{SSIM}\\
\midrule
1 Stage & 23.18 & 0.7920\\
2 Stage & 23.46 & 0.7956\\
\textbf{3 Stage} & \textbf{24.03} & \textbf{0.8030}\\
4 Stage & 23.72 & 0.7933\\
\bottomrule
\end{tabular}
\end{table}

\subsubsection{Robustness to Varying Angle Inaccuracy}
The core claim of our work is the model's robustness to uncertainty in the initial angle estimate. In Table~\ref{tab:robust_angle}, we compare our baseline model and the full model when applied to the test set under perfect angle information ($\sigma=0$) and under highly-noisy angles ($\sigma=5$). The results are informative. As expected, the performance of CAR-Net-Base mode degrades when high noise is introduced. In contrast, our fully integrated CAR-Net-ADM model demonstrate remarkable stability. Its performance in both noise-free and high-noise scenarios exhibits statistically equivalent results. This indicates the capability of the angle detection module can fully recover from initial angle errors of $\sigma=5$, restoring deblurring performance to an ideal noise-free state. This finding provides strong validation for the core contribution of our research.

\begin{table}[tbp]
\centering
\caption{Robustness to varying levels of initial angle noise}
\label{tab:robust_angle}
\begin{tabular}{c cc}
\toprule
\textbf{Method} & \textbf{$\sigma=0$ (No Noise)} & \textbf{$\sigma=5$ (High Noise)}\\
\midrule
CAR-Net-Base &23.94/0.8063 &23.83/0.8034\\
\textbf{CAR-Net-AD} & \textbf{24.03/0.8030} & \textbf{24.03/0.8030}\\
\bottomrule
\end{tabular}
\end{table}

\subsubsection{Loss function weight}
Our research revealed a critical interaction between Physics-based Consistency Loss ($w_{\mathrm{physics}}$) and the statistical complexity of training data. As shown in Table~\ref{tab:loss_ablation}, this relationship is demonstrated by comparing performance when training on our Simple Pattern versus our Real-World datasets. When training on statistically simple pattern datasets, a strong physical prior is crucial. A low weight ($w_{\mathrm{physics}}=0.1$), results in training failure, resulting in a poor PSNR of only 12.70 dB. Increasing the weight to 1.0 stabilized the training and improved the results to a more reasonable 19.04 dB.

In contrast, when training on the rich, statistically complex real-world datasets, the model can learn powerful natural image priors directly from the data. The optimal performance is archived under smaller weight ($w_{\mathrm{physics}}=0.1$), resulting in 24.03 dB by allowing the refinement network sufficient flexibility. Choosing a larger weight ($w_{\mathrm{physics}}=1.0$) slightly hurts the performance, resulting in 23.80 dB, as it prevents the model from learning details that go beyond the simple physics simulation.

This experiment on loss weights revealed a key insight: the optimal loss weighting is domain-dependent in our CAR-Net model. Simpler domains require stronger physical priors, while complex domains benefit from a more flexible, data-driven learning approach.

\begin{table}[tbp]
\centering
\caption{Impact of the physics-based consistency loss weight ($w_\mathrm{physics}$) during training on the Simple Pattern dataset}
\label{tab:loss_ablation}
\begin{tabular}{cc cc}
\toprule
\textbf{Training Data} & \textbf{Physics Loss} & \textbf{PSNR (dB)} & \textbf{SSIM}\\
\midrule
Simple Pattern & $w_\mathrm{physics}=0.1$ & 12.7 & 0.5644\\
Simple Pattern & \textbf{$w_\mathrm{physics}=1.0$} & 19.04 & 0.7326\\
Real World & $w_\mathrm{physics}=0.1$ & \textbf{24.03} & \textbf{0.8030}\\
Real World & \textbf{$w_\mathrm{physics}=1.0$} & 23.80 & 0.8022\\
\bottomrule
\end{tabular}
\end{table}

\subsubsection{Impact of Training Data}
We also conducted experiments to understand the impact of training data. Our experiments revealed two key insights regarding the composition of training data, with comparable results shown at Table~\ref{tab:training_data}. 
First, the statistical complexity of the training data is a fundamental prerequisite for a model to learn effective restoration functions. When training the full model with only Simple Pattern Datasets, we observed that the model performs extremely poorly, achieving PSNR 19.04 dB. This result strongly indicates that our model did not learn a general deblurring function but instead learned a strong natural image prior knowledge. The lack of complex statistical regularities in simple patterns leads the network from distinguishing real details from deconvolution artifacts. 

Second, training with angular noise is crucial for the performance of the angle detection model. When training the CAR-Net-AD model on real-world dataset using only perfect angles ($\sigma=0$), its ability to correct noise angles during testing is limited, with PSNR performance dropping to 23.67 dB under $\sigma=5$ noise. This confirms that ADM must be explicitly exposed to various deconvolution artifacts caused by noise angles during training to learn its correction function.

Therefore, we conclude that the optimal training strategy for our framework requires a dataset that is not only statistically rich (such as natural images) but also representative of the target problem (such as parameter uncertainty).

\begin{table}[tbp]
\centering
\caption{Impact of training data domain and noise level on performance. All models were evaluated on the Real-World test set ($\sigma=5$)}
\label{tab:training_data}
\begin{tabular}{cc c}
\toprule
\textbf{Method} & \textbf{Training Data} & \textbf{Result}\\
\midrule
CAR-Net-Base & Simple ($\sigma=5$)& 20.50/0.7543\\
CAR-Net-Base & Real ($\sigma=0$) & 23.64/0.7962\\
CAR-Net-Base & Real ($\sigma=5$) & 23.83/0.8034\\
CAR-Net-AD & Simple ($\sigma=5$) & 19.04/0.7326\\
CAR-Net-AD & Real ($\sigma=0$) & 23.67/0.7952\\
\textbf{CAR-Net-AD} & \textbf{Real ($\sigma=5$)} & \textbf{24.03/0.8030}\\

\bottomrule
\end{tabular}
\end{table}

\subsection{Comparison with State-of-the-Art}
We compare the performance of CAR-Net with state-of-the-art model PCS-RMD \cite{qin2025} by using its pre-trained model officially released in the repository of PCS-RMD. The comparison is performed on Real-World test set which was mentioned in Section~\ref{Sec:Experiment_setup} under angle uncertainty of $\sigma=5$. 

It is important to note the key differences in training data distribution, as this comparison primarily test the generalization of both models. As state in their paper, PCS-RMD model was trained on a narrow and specific range of blur angles $[5.04\degree,10.08\degree]$. In contrast, our model was trained on a broader and more diverse range $[1\degree,40\degree]$ to enhance its robustness across various degrees of blur. 
In addition, the blur synthesis method differ: their approach is average a series of rotated image, while our data was generated using GIMP.

Thus, this comparison tests the robustness of each model against a broader spectrum of unseen blur condition. As shown in Table~\ref{tab:main_results}, the results demonstrate that our proposed frameworks show a better performance. The full model CAR-Net-AD achieves the highest PSNR and SSIM scores, validating the effectiveness of this architecture and its strong generalization capabilities.

\begin{table}[tbp]
\centering
\caption{Comparison with the state-of-the-art method on the Real-World test set}
\label{tab:main_results}
\begin{tabular}{c cc}
\toprule
\textbf{Method} & \textbf{PSNR (dB)} & \textbf{SSIM}\\
\midrule
PCS-RMD & 18.94 & 0.6560\\
CAR-Net-Base (Ours) & 23.83 & 0.8034\\
\textbf{CAR-Net-AD (Ours)} & \textbf{24.03} & \textbf{0.8030}\\
\bottomrule
\end{tabular}
\end{table}

\subsection{Model Complexity and Efficiency}
We analyzed each model's practical efficiency across two dimensions: parameter count and inference speed. Table~\ref{tab:complexity} presents comparison results of our models against the current state-of-the-art baseline model~\cite{qin2025}. The parameter count was determined by summing the number of elements for each tensor within the model's state dictionary and all inference times were bench-marked on a single NVIDIA RTX 3080ti GPU for 320x320 resolution images.
The results highlight the significant advantages of our proposed CAR-Net architecture. The baseline model CAR-Net-Base is exceptionally lightweight, containing only 0.31 million parameters, over five times smaller than the PCS-RMD baseline. This compact size translates directly to a remarkable inference speed of just 7.2ms, making it over 5.4 times faster than PCS-RMD. 
Our full model CAR-Net-AD, incorporating an innovative angle correction mechanism, adds approximately 1M parameters. However, it remains more compact than the PCS-RMD baseline. Crucially, our CAR-Net-AD architecture is more efficient, with processing time around 9ms. As a model, CAR-Net-AD not only achieves greater accuracy and robustness but also delivers inference speeds 4.3 times faster than the current state-of-the-art model. This demonstrates that our proposed framework can achieve significant performance without sacrificing computational efficiency, making it highly suitable for practical applications.

\begin{table}[tbp]
\centering
\caption{Comparison of model complexity and efficiency}
\label{tab:complexity}
\begin{tabular}{c cc}
\toprule
\textbf{Method} & \textbf{No. of Params ($10^6$)} & \textbf{Inference Time (ms)}\\
\midrule
PCS-RMD \cite{qin2025} & 1.65 & 39.2 \\
CAR-Net-Base (Ours) & \textbf{0.31} & \textbf{7.2} \\
CAR-Net-AD (Ours) & 1.38 & 9.0 \\
\bottomrule
\end{tabular}
\end{table}

\subsection{Qualitative Results}
\begin{figure*}[tbp]
    \centering
    \begin{subfigure}{0.19\textwidth}
        \includegraphics[width=\linewidth]{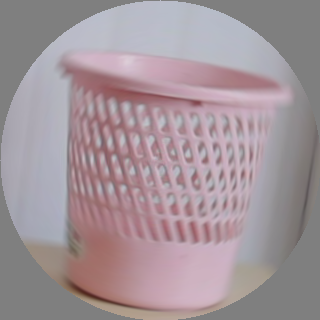}
        \caption{Blurred}
        \label{fig:blurred}
    \end{subfigure}
    \hfill
    \begin{subfigure}{0.19\textwidth}
        \includegraphics[width=\linewidth]{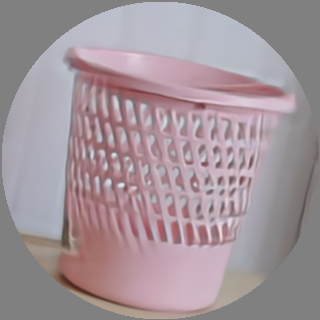}
        \caption{PCS\_RMD/23.38}
        \label{fig:pcs_rmd}
    \end{subfigure}
    \hfill
    \begin{subfigure}{0.19\textwidth}
        \includegraphics[width=\linewidth]{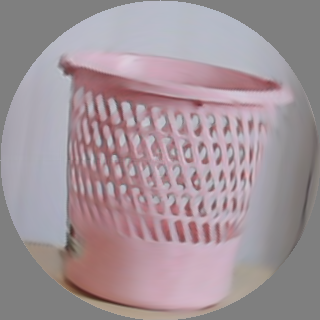}
        \caption{CAR-Net-Base/31.23}
        \label{fig:ours1}
    \end{subfigure}
    \hfill
    \begin{subfigure}{0.19\textwidth}
        \includegraphics[width=\linewidth]{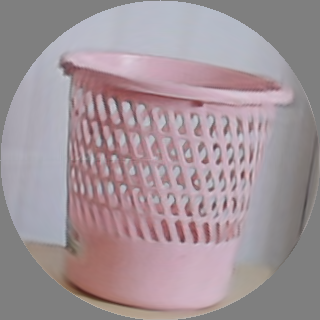}
        \caption{\textbf{CAR-Net-AD/32.42}}
        \label{fig:ours2}
    \end{subfigure}
    \hfill
    \begin{subfigure}{0.19\textwidth}
        \includegraphics[width=\linewidth]{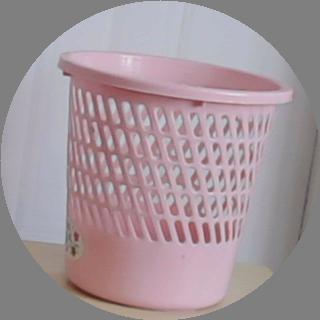}
        \caption{Ground Truth}
        \label{fig:ground_truth}
    \end{subfigure}

    \caption{A visual comparison of deblurring results on an image from the Real-World test set. 
    (Noise uncertainty level $\sigma=5$, ground truth angle $\theta_{\mathrm{GT}}=5.41\degree$, noisy initial angle $\theta_{\mathrm{initial}}=8.98\degree$).}
    \label{fig:deblurring_comparison}
\end{figure*}

To provide a visual assessment of our model's performance, we present a qualitative comparison in Fig~\ref{fig:deblurring_comparison}. The figure displays examples from our Real-World test set, comparing our model against the state-of-the-art PCS-RMD baseline.

As shown, while all methods successfully eliminate most rotational blur, significant differences exist in restoration quality. The PCS-RMD model's output shows a significant improvement over the blurred input but insufficient sharpness, indicates residual ringing artifacts especially at right upper corner of the image, particularly noticeable in the complex grid structure of the basket. Our baseline model CAR-Net-Base produces shaper and cleaner result, validating the effectiveness of our progressive refinement module in artifact suppression.
The full model CAR-Net-ADM demonstrates particularly pronounced advantages, especially in the right region. By first identifying the correct blur angle, this model provides the refinement network with an optimal initial state. This enables sharper edges and more faithful reproduction of fine details like the plastic mesh. As a result, the reconstructed image closely matches the real image and appears to offer the best visual experience.

\section{Discussion}
\label{Sec:Discussion}
\subsection{The Efficacy of a Hybrid, Modular Approach}
A key finding is that our relatively compact model outperforms larger PCS-RMD baseline model. We attribute this to our hybrid modular design: the non-learned module Inversion Module, based on explicit physical models, handles the important role of global deconvolution. This allows our learned refinement network (refinement stages) to focus on the more constrained and simple tasks of artifact removal and texture synthesis, rather than learning the entire complex inverse problem from scratch. This division of responsibilities results in a more efficient and effective network that not only resists overfitting but also demonstrates strong generalization capabilities.

\subsection{Comparison to State-of-the-Art Approaches}
While our CAR-Net framework demonstrates superior performance in the challenging semi-blind scenario, it is important to acknowledge the distinct strengths of the PCS-RMD approach. A key aspect of their framework is interpretability, which stems from its modular design targeting specific, known problems.
For instance, their dedicated module for correcting Cartesian-to-Polar (CPT) and Polar-to-Cartesian (PCT) transformation artifacts represents a fundamental and powerful solution to the core challenge in PCS-based methods.
By identifying a clear source of error---resampling and interpolation---and designing a specific network to address it, their approach is not only effective but also highly flexible in directly targeting known errors. This makes it highly adaptable to other applications where coordinate transformation is used. 
In contrast, our work prioritizes a different challenge: decoupling parameter correction from image refinement in a lightweight, cascaded architecture. This strategy proves highly effective for robustly solving the parameter uncertainty problem.

\subsection{Fidelity of the Blur Simulation and the 2D Assumption}
It is important to know that there is an inherent domain gap between our synthesized training data and real-world physical rotational motion blur. This gap exists on two levels: the blur synthesis process itself, and the underlying assumption of two-dimensional image formation.
First, our data is generated using GIMP's circular motion blur filter. This filter applies 2D convolution algorithm to simulate rotational motion blur, based on the assumption of a perfect camera sensor and a fixed rotational center point. While the filter effectively produces controllable blur effects, it fails to simulate complex real-world camera effects such as varying angular velocity, lens distortion during motion, and the interaction between motion and different camera shutter mechanisms. These physical details generate blur characteristics that differ from a pure algorithmic simulation.
Second, our entire model operates on a 2D image-space assumption. We model the blur as a 2D transformation of a single, static sharp image. In reality, a camera captures a projection of a 3D scene over time. As the camera rotates, previously masked parts of the scene, such as the side of an object invisible in the initial pose. These newly revealed scene parts (the deocclusion regions) do not exist in any single baseline clear image, posing a fundamental challenge to 2D deblurring models. Our 2D simulation cannot capture these complex 3D effects as the information from newly revealed region (deocclusions)is entirely absent from training data, meaning model are not trained to reconstruct them. This in turn implies that models may fail to perfectly match certain real-world scenarios. Addressing the issue may require more sophisticated data acquisition methods (such as light fields or high-speed video) and novel network architectures designed to handle the understanding of 3D scenes.

\subsection{Limitations}
While our framework demonstrate strong performance, several limitations remain and open up directions for future research:
\begin{itemize}
    \item \textbf{Semi-blind assumption:} Our model is not fully blind. It requires a reasonably close initial estimate of the blur angle. If the initial guess is significantly off or falls outside the distribution observed during training, performance may degrade significantly.
    \item \textbf{Fixed Rotational Center:} The current model assumes a known and fixed rotational center. In many real-world scenarios, the rotational center is unknown and may vary.
    \item \textbf{PCS Transformation Artifacts:} Like all PCS-based methods, the final output exhibits undefined image corners. Although we mitigated this issue in evaluation by using circular ROIs, it remains a practical limitation for full-frame image restoration.
\end{itemize}

\subsection{Implications}
The success of our work has two implications. For a practical standpoint, the model's high accuracy and computational efficiency makes it a useful solution for computational photography, drone/satellite image enhancement, and forensic analysis. Methodologically, the good performance of the angle detection module validates a powerful principle: employing a specialized lightweight neural network to predict and correct errors of a physical model parameter represents an efficient strategy for hybrid image restoration. This corrective cascade approach could be adapted to other problems with parameter uncertainty. 

\subsection{Future Work}
There are several promising research directions for future research. One of those directions is to extend our framework to the fully blind setting, training the network to estimate both rotational center and angle collaboratively from scratch. We also plan to explore the model's transferability to the case of real-world blurred images through fine-tuning. In this regard, we aim to investigate a more complex motion model such as non-uniform rotational velocities and address the 3D-to-2D projection domain gap. Finally, it is possible to extend our proposed cascade to the field of 3D rotational deblurring with potential applications in medical imaging (CT/MRI) or video restoration scenarios.

\section{Conclusion}
In this paper, we introduced CAR-Net, a novel multi-stage framework for semi-blind rotational motion deblurring in Polar Coordinate System. We demonstrated two configuration schemes: the baseline hybrid model combines frequency-domain inversion with progressive spatial refinement, while the full model integrates an innovative angle detection module. This module learns to correct the initial blur angle error, making the recovery process robust against parameter uncertainties.
We conducted extensive experiments and ablation studies to validate the effectiveness of our design, evaluating the contribution of each component to the final performance. Our final model, CAR-Net-AD, outperforms the state-of-the-art method in both quantitative metrics and subjective evaluation over visual quality, while achieving substantial computational efficiency gains. Our work demonstrates the powerful capabilities of a modular, physically guided cascaded architecture that decouples parameter correction from image refinement.

\bibliographystyle{IEEEtran}
\bibliography{references}

\end{document}